# Diagnosis of Covid-19 Via Patient Breath Data Using Artificial Intelligence

Özge Doğuç [1*], Gökhan Silahtaroğlu [1], Zehra Nur Canbolat [1], Kailash Hambarde [2], Ahmet Alperen Yiğitbaşı [1], Hasan Gökay [1], Mesut Yılmaz [1]

[1] Department of Management Information Systems, Istanbul Medipol University, Istanbul, Turkey.

[2] School of Computational Sciences, SRTM University Nanded, Nanded, India.

**Abstract**

Using machine learning algorithms for the rapid diagnosis and detection of the COVID-19 pandemic and isolating the patients from crowded environments are very important to controlling the epidemic. This study aims to develop a point-of-care testing (POCT) system that can detect COVID-19 by detecting volatile organic compounds (VOCs) in a patient's exhaled breath using the Gradient Boosted Trees Learner Algorithm. 294 breath samples were collected from 142 patients at Istanbul Medipol Mega Hospital between December 2020 and March 2021. 84 cases out of 142 resulted in negatives, and 58 cases resulted in positives. All these breath samples have been converted into numeric values through five air sensors. 10% of the data have been used for the validation of the model, while 75% of the test data have been used for training an AI model to predict the coronavirus presence. 25% have been used for testing. The SMOTE oversampling method was used to increase the training set size and reduce the imbalance of negative and positive classes in training and test data. Different machine learning algorithms have also been tried to develop the e-nose model. The test results have suggested that the Gradient Boosting algorithm created the best model. The Gradient Boosting model provides 95% recall when predicting COVID-19 positive patients and 96% accuracy when predicting COVID-19 negative patients.

## 1- Introduction

The coronavirus (SARS-CoV-2, 2019-nCoV), which caused the COVID-19 epidemic, spread significantly around the world and caused many casualties [1]. The disease initially spread rapidly in Southeast Asia and Europe, and as a result, it was declared a global pandemic by the World Health Organization (WHO) on March 11, 2020 (WHO briefing).

The SARS-CoV-2 virus is highly contagious, so tests that can quickly and accurately identify patients with SARS-CoV-2 infection have been needed throughout the pandemic. Many patients with SARS-CoV-2 infection didn't show any symptoms, and they had come into contact with other people before they were diagnosed [1-3]. Therefore, there has been a need for a rapid, inexpensive, and highly accurate point-of-care testing (POCT) method for the timely isolation of infected cases and effective monitoring of potential new cases. [4]. With high accuracy, POCTs can have a greater impact than RT-PCR, which requires technical expertise and laboratory capacity for disease detection, occasionally results in false-negative results [5, 6]. Therefore, countries employ different measures to protect public safety, such as requiring vaccination cards, rapid tests, etc., none of which can provide an accurate representation of the COVID-19 infection.

---

* **CONTACT**: odoguc@medipol.edu.tr

The goal of this study is to develop a POCT system that can detect COVID-19 by accurately decomposing volatile organic compounds (VOCs) in a patient's breath. There are studies in the literature that show that deep learning can be effectively used in the detection and diagnosis of COVID-19, particularly through radiology modalities [5–7]. For this purpose, a hand-held electronic nose (e-nose) device is designed and built. The device contains a tube that patients blow into, and it can accurately detect the existence of a SARS-CoV-2 infection in just a few seconds. Operating the device doesn't require any special training, and it is designed to be used in public areas such as stadiums, airports, restaurants, and shopping malls.

## 2- Material and Method

### 2-1- E-Nose Structure

The e-noses that have been built for this study employ five different gas sensors: MQ2, MQ3, MQ7, MQ8, and MQ135. Each sensor is sensitive to a different gas compound in human breath and detects the presence of the gas within a range of 0-1000 ppm (parts per million) [8].

The sensors used in the e-noses can be summarized as below:

- MQ2 is a combustible gas sensor. It has high sensitivity to LPG, Propane, Methane, and other combustible gases.
- MQ3 is a cork gas sensor suitable for alcohol, gasoline, CH4, Hexane, LPG and carbon monoxide detection.
- MQ7 is a sensor that is very sensitive to carbon monoxide.
- MQ8 sensor is used for detecting high concentrations of the Hydrogen gas.
- The MQ135 air quality sensor can detect the presence of many gases, especially NH3, benzene, alcohol and carbon dioxide [9].

### 2-2- Application

For this study, a handheld e-nose device is built for data collection and breath analysis, as shown in Figure 1. E-noses were used for coronavirus detection in 142 patient cases at Medipol Mega Hospital between December 2020 and March 2021. For each patient, two or all the following methods have been used for coronavirus testing:

- Breath analysis with e-noses in specialized cabins;
- Nasal and throat swabs;
- PCR tests.

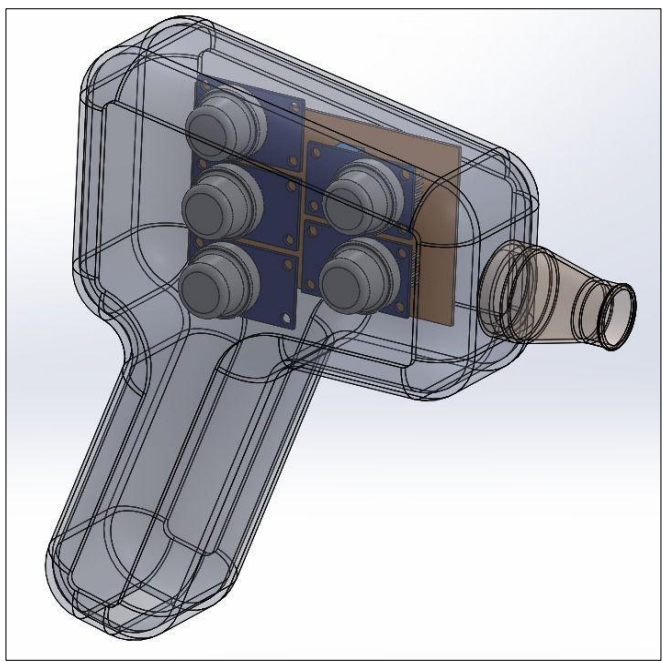

**Figure 1. E-nose device model**

Out of 292 breath samples collected from 142 patients 84 cases resulted negative, and 58 cases resulted positive. Collected data is stored in a database, which is then used for creating an accurate artificial intelligence (AI) model for disease detection. Figure 2 presents the workflow of the approach used in this study.

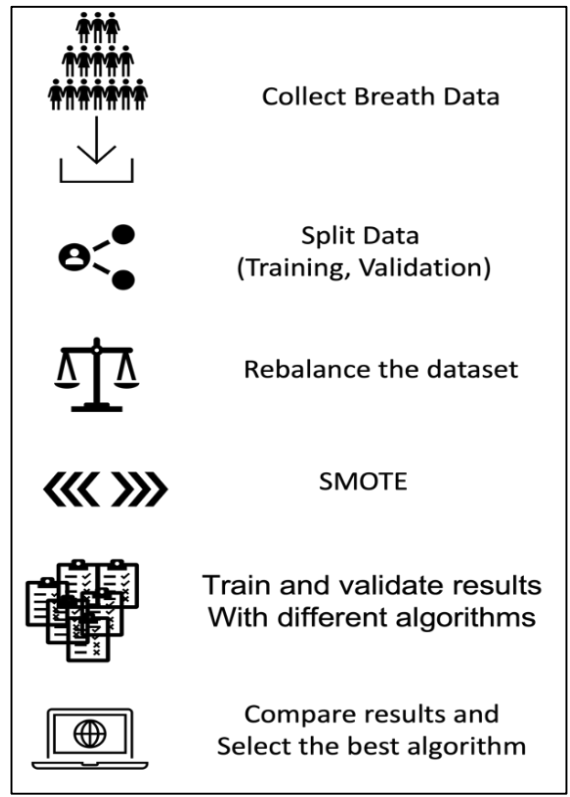

**Figure 2.** Workflow of the study approach

*2-2-1- Data Preparation*

While PCR test result data based on nasal and throat swabs is binary (positive or negative), sensors in the e-nose generate non-discrete numeric values. This allows the analysis of results from combinations of multiple sensors. For example, while results from the MQ2 sensor can have a strong impact on the coronavirus test results, a combination of MQ2 and MQ3 together may provide a stronger association with the coronavirus test results. For this purpose, this study also considers the following metrics and their squares:

- MQ2 - MQ3;
- MQ2 - MQ7;
- MQ2 - MQ8;
- MQ2 - MQ135;
- MQ3 - MQ7;
- MQ3 - MQ8;
- MQ3 - MQ135;
- MQ7 - MQ8;
- MQ7 - MQ135;
- MQ8 - MQ135.

The resulting dataset used in this study has 27 attributes, and 294 records. 90% of this data (264 records) is used for training an AI model to predict the coronavirus presence, and the remaining 10% (30 records) is used for validating the results. Out of the 264 records, 75% were used for training the model and 25% were used for testing. Only after successful testing results were achieved, the model was used for validating the result with the 10% of the data. Figure 3 shows the details of the data preparation step.

Standard classifiers give biased results in the direction of the larger subset when the dataset is unevenly distributed. The dataset used in this study is also unstable with a 4:1 ratio, and standard users may give incorrect results. So, before training, a model must address this issue. Figure 4 shows two types of data sampling: over sampling and under sampling. In under sampling algorithm, majority class blue points are reduced to the same size as the minority class red data points. In over sampling minority class, the red data points increased to the same size of the majority class blue data points [10, 11].

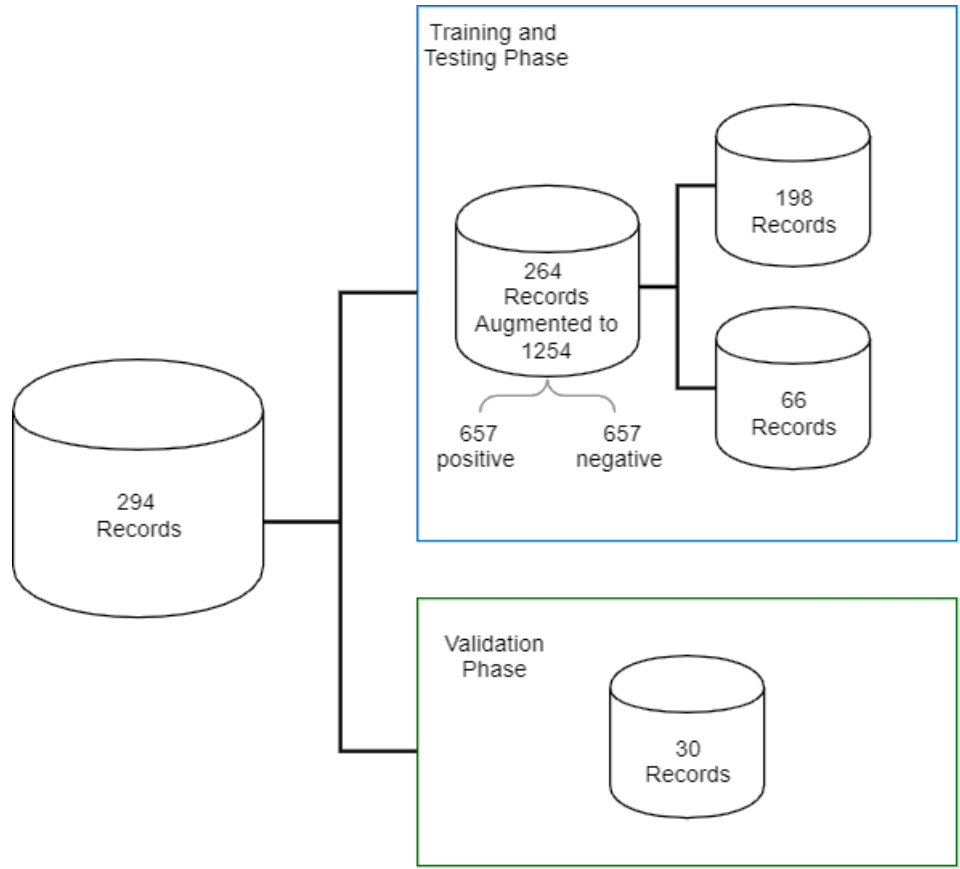

**Figure 3. Data Preparation Process**

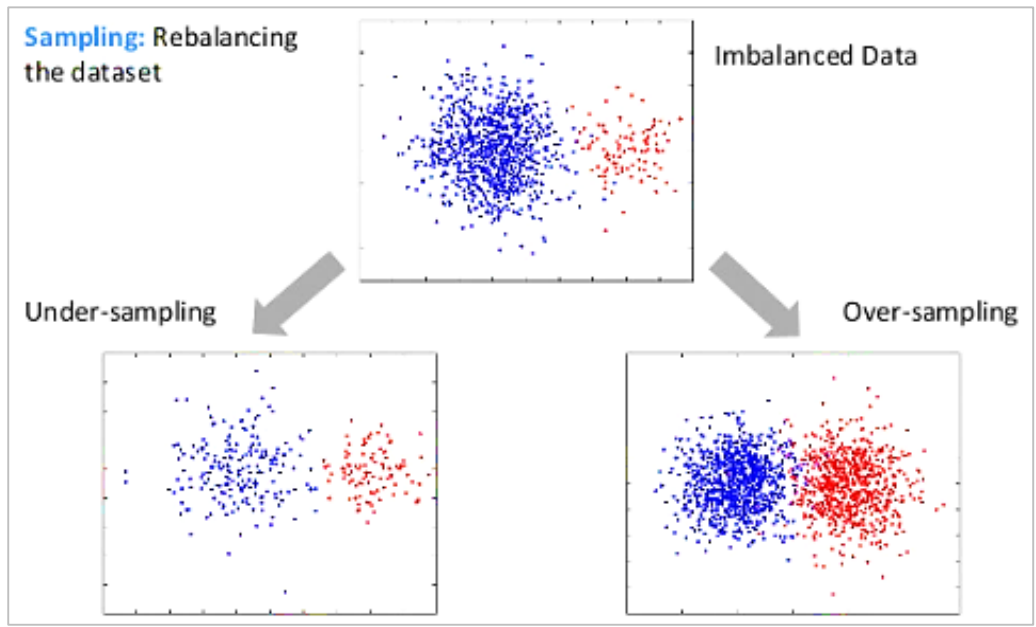

**Figure 4. Rebalancing the dataset**

The next step used in data preparation is to synthetically increase the training set size and reduce the imbalance of negative and positive class sizes. Increased training set size offers more accurate results and balancing the data set reduces overfitting (i.e., learning majority cases only). For this purpose, the SMOTE (Synthetic Minority Oversampling Technique) oversampling method was used. SMOTE is a popular method that generates synthetic data for the minority data classes. Because majority of the collected breath data is Covid-negative (i.e., doesn't contain any trace of the SARS-CoV-2 virus), SMOTE helped balancing the negative and positive cases in the dataset. We used SMOTE data resampling techniques to solve the problem. SMOTE is a sampling algorithm that implements the k-nearest neighbor (KNN) algorithm approach. The algorithm selects the K nearest neighbors, combines them, and generates synthetic data as a result. (Figure 5).

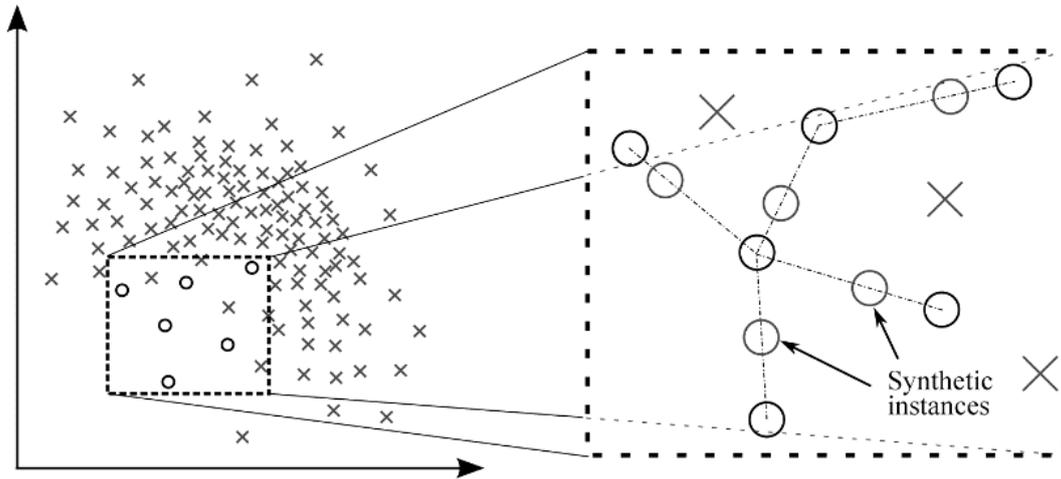

**Figure 5.** Over Sampling Algorithms based on SMOTE [11]

After the balancing phase, data was augmented to achieve a better training. After augmentation the data set size increased to 1,254 records. Table 1 shows the number of records and class in the dataset which has been used in the study.

**Table 1.** Information about the data used

| Sr | Dataset name | Total Dataset | Majority class (negative cases) | Minority class (positive cases) | Imbalance ratio |
|---|---|---|---|---|---|
| 1 | Original Data | 294 | 233 | 61 | 4:1 |
| 2 | Balanced Data (90% of the original) | 418 | 209 | 209 | 1:1 |
| 3 | Augmented Data for Training | 1,254 | 627 | 627 | 1:1 |
| 4 | Data for Validation (No augmentation or SMOTE) | 30 | 24 | 6 | 4:1 |

After the data has been balanced, we have trained a gradient boosting algorithm to check the efficiency of the model. The impact of balancing the dataset can be seen in the initial test results that are given in Table 2. For model evaluation, we used recall and precision values from the confusion matrix. The original (imbalanced) dataset was initially used to train an imbalanced model by using the gradient boosting algorithm. This dataset was imbalanced with ratio 4:1 meaning, for every 4 negative patient there are 1 positive cases. This model provides 68% precision, namely, out of 100 COVID-negative patients, the unbalanced model predicts 68 patients as negative (true negative) and 32 patients as positive (false negative). Also, this model results in a 70% recall, which means that out of 100 Covid-19 positive patients, the imbalanced model predicts 70 patients as Covid-19 positive (true positive) and 30 patients as negative (false positive). Table 2 shows that all precision, recall, and accuracy results have improved greatly after balancing the dataset and retraining the model. A balanced model predicts 95% recall, 96% precision, and 96% accuracy.

**Table 2.** Evaluation of ML Model both for original and balanced datasets

| Sr | Dataset name | Precision | Recall | Accuracy |
|---|---|---|---|---|
| 1 | Original Data | 68% | 70% | 68% |
| 2 | Balanced Data | 96% | 95% | 96% |

*2-2-2- Gradient Boosted Trees Learner Algorithm*

This study uses the KNIME platform [12] to visualize data and create learning models. The learning algorithm that was chosen for this study is Gradient Boosted Trees. In the case of the gradient-assisted decision tree algorithm, the weak learners are the decision trees, and they are prone to the problem of overfitting. To reduce this risk, a model combining multiple decision trees is used in this study. Random forests use a method called bagging to combine many decision trees into a single tree. At each iteration, the random forests randomly pick any number of features, and create decision trees accordingly. The results of the decision trees are aggregated based on the voting principle. Random selection of features solves that overfitting problem that is present in decision trees [13].

The gradient boosted trees learner algorithm can be described as follows:

On a dataset (x, y) with x features and y targets, the loss function $L$ can be calculated as the Squared Residuals as follows:

$$L = \frac{1}{2}(Obs - Pred)^2 \tag{1}$$

where *Obs* and *Pred* show the observed and predicted values respectively. The *L* function is differentiable:

$$\frac{\partial}{\partial Pred} L = -1 \, x \, (Obs - Pred) \tag{2}$$

- Initialization:

$$F_0(x) = argmin_\gamma \sum_i L(y_i, \gamma) \tag{3}$$

The algorithm tries to choose the best prediction by minimizing the *L* function (squared residuals). Deriving the optimal value for the class would provide predictions that will weigh the average of the samples.

$$\frac{\partial}{\partial \gamma} \sum_i L(y_i, \gamma) = -(y_1 - \gamma) - (y_2 - \gamma) - (y_3 - \gamma) \ldots = 0 \tag{4}$$

$$\sum_i y_i - n * \gamma = 0 \tag{5}$$

$$\gamma = \frac{\sum_i y_i}{n} = \bar{y} \tag{6}$$

- For t = 1 to M (maximum number of trees)

  o Calculate the pseudo-residuals at every iteration. This derivative is called the *Gradient*

$$r_{it} = -\frac{\delta L(y_i, F(x_i))}{\delta F(x_i)} = -(-1 \times (Obs - F_{t-1}(x)) = (Obs - F_{t-1}(x)) = (Obs - Pred) \tag{7}$$

  o Fit a regression tree to the rim values and create terminal regions $R_{jt}$ for $j = 1, \ldots, J_t$ (create the leaves of the tree). At that point, the output value of each leaf still needs to be computed.

  o For each leaf $j = 1\ldots J_m$, compute the output value that minimizes the sum of squared residuals (SSR). Outputs of all samples stored in a certain leaf will be predicted.

$$\gamma_{jt} = argmin_\gamma \sum_{x_i \in R_{ij}} L(y_i, F_{t-1} + \gamma) \tag{8}$$

  o Make a new prediction for each sample by updating, according to a learning rate $l_r \in (0,1)$:

$$F_t(x) = F_{t-1}(x) + l_r \, x \, \sum_j \gamma_{jt} I(x \in R_{jt}) \tag{9}$$

The new value is computed by summing the previous prediction and all the predictions into which the sample falls [14].

## 3- Results and Discussion

In this study, besides the gradient boosting Machine Learning (ML) algorithm, other ML algorithms such as logistic regression, gradient boosting, random forest support vector machine, KNN, decision tree, and Naïve Bayes have also been used for comparison. Nevertheless, as seen in Table 3, gradient boosting surpasses all others in terms of precision, recall, and accuracy values. However, it is well known that, depending on the composition of the dataset, other algorithms may also achieve good predictions, but in our case, gradient boosting provided the best results.

Table 3. **Performances of all ML algorithms**

| Sr | Algorithm | Precision | Recall | Accuracy |
|---|---|---|---|---|
| 1 | Logistic Regression | 0.91 | 0.84% | 0.89 |
| **2** | **Gradient Boosting** | **0.96%** | **0.95%** | **0.96%** |
| 3 | Random Forest | 0.93% | 0.89 | 0.92 |
| 4 | Support Vector Machine | 0.86 | 0.75 | 0.72 |
| 5 | KNN | 0.94 | 0.85 | 0.91 |
| 6 | Decision Tree | 0.69 | 0.71 | 0.73 |
| 7 | Naïve Bayes | 0.77 | 0.82 | 0.93 |

Finally, performance of Gradient Boosting has been evaluated, as seen in Table 4. Statistics like specificity, positive predictive value (PPV), negative predictive value (NPV) and others besides receiver operating characteristics (ROC) curve suggest that model can be used for predictions. That means e-nose may be used in place of PCR tests.

Table 4. **Detailed performance analysis of Gradient Boosting**

| Results | Testing | | Validation | |
|---|---|---|---|---|
| Statistic | Value | 95% CI | Value | 95% CI |
| Sensitivity | 95.62% | 91.19% to 98.22% | 80.00% | 51.91% to 95.67% |
| Specificity | 96.18% | 91.87% to 98.58% | 96.43% | 87.69% to 99.56% |
| Positive Likelihood Ratio | 25.02 | 11.41 to 54.88 | 22.4 | 5.61 to 89.42 |
| Negative Likelihood Ratio | 0.05 | 0.02 to 0.09 | 0.21 | 0.08 to 0.57 |
| Disease prevalence (*) | 50.47% | 44.83% to 56.11% | 21.13% | 12.33% to 32.44% |
| Positive Predictive Value (*) | 96.23% | 92.08% to 98.24% | 85.71% | 60.05% to 95.99% |
| Negative Predictive Value (*) | 95.57% | 91.26% to 97.80% | 94.74% | 86.73% to 98.02% |
| Accuracy (*) | 95.90% | 93.09% to 97.80% | 92.96% | 84.33% to 97.67% |

Table 4 shows that the trained model can very accurately detect true negatives (specificity). In other words, if the e-nose decides that someone doesn't have the SARS-CoV-2 virus, the probability that this person is infected is less than 4%. The same table also shows that both PPV and NPV are higher than 85%, and accuracy is almost 93%. ROC curves for both training and validation are given in Figures 6 and 7, respectively. The ROC curves are plotted against the PCR test results, and the area under the curve (AUC) shows that the trained models have the same separation level as the PCR tests.

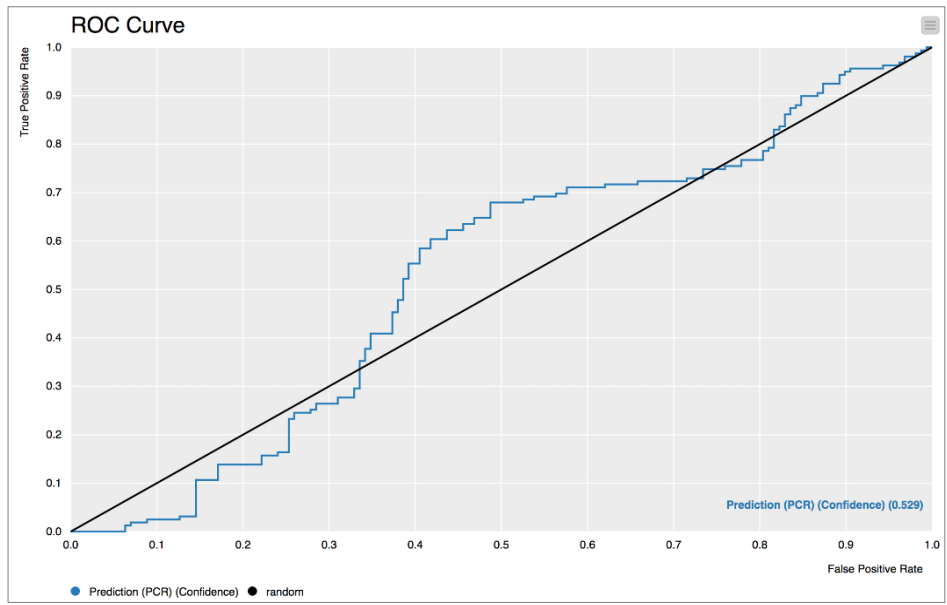

Figure 6. **Testing ROC**

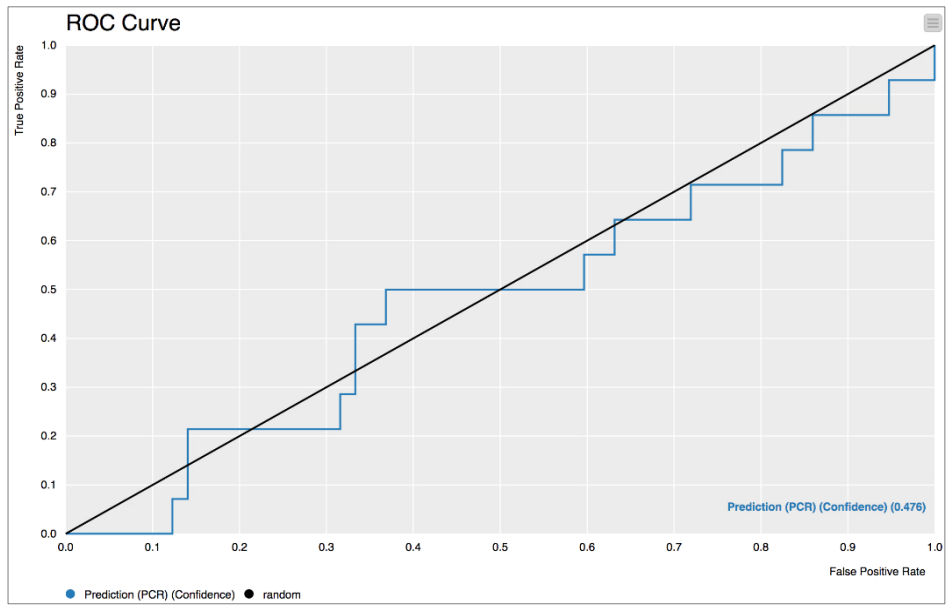

Figure 7. **Validation ROC**

### 3-1- E-Nose Device

For this study, a handheld device for breath analysis is built as seen in Figure 8 (patent pending). The device is equipped with the gas sensors that were listed in the Introduction section. It allows the patients to blow through a reusable silicone tube and generates a diagnostic output for COVID-19 in a few seconds. The e-nose device also has wireless and Bluetooth capabilities for easy data transfer to computers. Also, newly trained models can be easily loaded onto the device through its communication port.

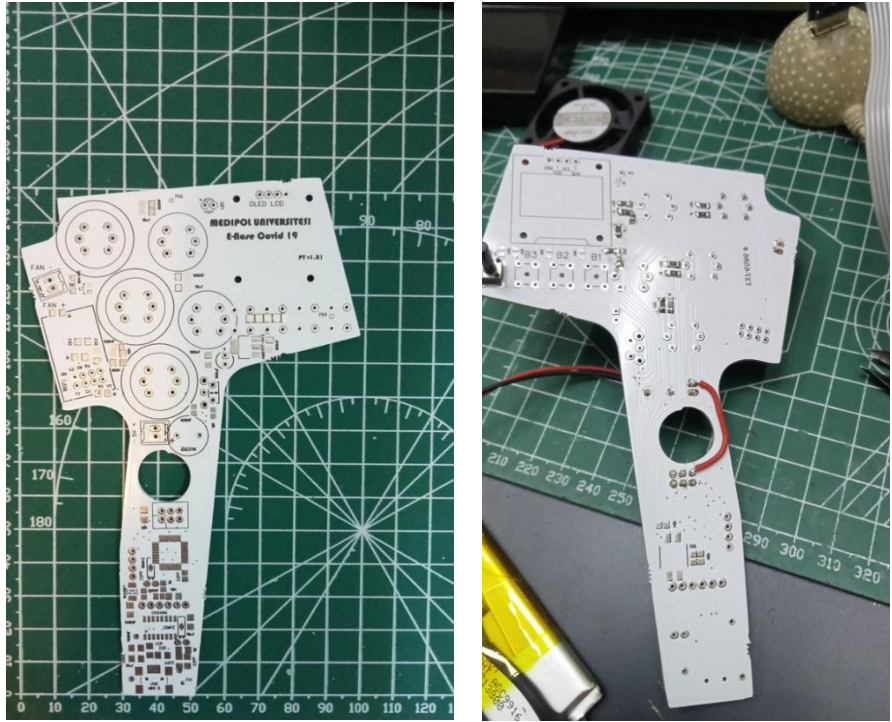

**Figure 8. The E-Nose Device**

## 4- Conclusion

Gradient boosting is one of the popular artificial intelligence / machine learning algorithms used in the literature. In this study, we used Gradient Boosting AI model on artificially balanced and augmented breath data taken from 142 patients through e-noses that were also designed and created for this study. All the patients who participated in the study either showed symptoms of COVID-19 in that way or another.

Data set have been divided in two sets as 10% validation and 90% for AI training. The AI training dataset has been divided into two portions: 75% training and 25% testing of the results. The AI training data has been balanced in terms of the class variable (i.e., Covid Positive or Negative) and augmented with the KNN algorithm. In fact, the study analyzes different AI algorithms' such as SMOTE, KNN, and Gradient Boosting AI algorithms, for their performance with the collected breath data. The class variable is derived from PCR test results; therefore, the study relies on PCR tests. Our results show that the e-noses can predict COVID-19 with 96% accuracy with respect to the PCR test results. Contrary to the PCR tests that require at least a few hours for the results, the e-noses can decide the COVID-19 outcome in a matter of seconds. Therefore, the e-noses can be used in place of the PCR tests as a quick and cheap alternative.

## 5- Declarations

### 5-1- Author Contributions

### 5-2- Data Availability Statement

The data presented in this study are available on request from the corresponding author.

### 5-3- Funding

The authors received no financial support for the research, authorship, and/or publication of this article.

### 5-4- Institutional Review Board Statement

The study was conducted in accordance with the Declaration of Helsinki and approved by the Ethics Committee of the Istanbul Medipol University (protocol code E-10840098-772.02-1168 on 21/03/2021).

### 5-5- Informed Consent Statement

Not applicable.

### 5-6- Conflicts of Interest

The authors declare that there is no conflict of interests regarding the publication of this manuscript. In addition, the ethical issues, including plagiarism, informed consent, misconduct, data fabrication and/or falsification, double publication and/or submission, and redundancies have been completely observed by the authors.

## 6- References

[1] Marcel, S., Christian, A. L., Richard, N., Silvia, S., Emma, H., Jacques, F., Marcel, Z., Gabriela, S., Manuel, B., Annelies, W. S., Isabella, E., Matthias, E., & Nicola, L. (2020). COVID-19 epidemic in Switzerland: On the importance of testing, contact tracing and isolation. Swiss Medical Weekly, 150(1112). doi:10.4414/smw.2020.20225.

[2] Cheng, H. Y., Jian, S. W., Liu, D. P., Ng, T. C., Huang, W. T., & Lin, H. H. (2020). Contact Tracing Assessment of COVID-19 Transmission Dynamics in Taiwan and Risk at Different Exposure Periods before and after Symptom Onset. JAMA Internal Medicine, 180(9), 1156–1163. doi:10.1001/jamainternmed.2020.2020.

[3] Kucharski, A. J., Klepac, P., Conlan, A. J. K., Kissler, S. M., Tang, M. L., Fry, H., Gog, J. R., Edmunds, W. J., Emery, J. C., Medley, G., Munday, J. D., Russell, T. W., Leclerc, Q. J., Diamond, C., Procter, S. R., Gimma, A., Sun, F. Y., Gibbs, H. P., Rosello, A., … Simons, D. (2020). Effectiveness of isolation, testing, contact tracing, and physical distancing on reducing transmission of SARS-CoV-2 in different settings: a mathematical modelling study. The Lancet Infectious Diseases, 20(10), 1151–1160. doi:10.1016/S1473-3099(20)30457-6.

[4] Hellewell, J., Abbott, S., Gimma, A., Bosse, N. I., Jarvis, C. I., Russell, T. W., Munday, J. D., Kucharski, A. J., Edmunds, W. J., Funk, S., Eggo, R. M., Sun, F., Flasche, S., Quilty, B. J., Davies, N., Liu, Y., Clifford, S., Klepac, P., Jit, M., … van Zandvoort, K. (2020). Feasibility of controlling COVID-19 outbreaks by isolation of cases and contacts. The Lancet Global Health, 8(4), e488–e496. doi:10.1016/s2214-109x(20)30074-7.

[5] Wang, W., Xu, Y., Gao, R., Lu, R., Han, K., Wu, G., & Tan, W. (2020). Detection of SARS-CoV-2 in Different Types of Clinical Specimens. JAMA - Journal of the American Medical Association, 323(18), 1843–1844. doi:10.1001/jama.2020.3786.

[6] Turkoglu, M. (2021). COVID-19 Detection System Using Chest CT Images and Multiple Kernels-Extreme Learning Machine Based on Deep Neural Network. Irbm, 42(4), 207–214. doi:10.1016/j.irbm.2021.01.004.

[7] Ghaderzadeh, M., & Asadi, F. (2021). Deep Learning in the Detection and Diagnosis of COVID-19 Using Radiology Modalities: A Systematic Review. Journal of Healthcare Engineering, 2021. doi:10.1155/2021/6677314.

[8] Visvam Devadoss Ambeth, K. (2016). Human security from death defying gases using an intelligent sensor system. Sensing and Bio-Sensing Research, 7, 107–114. doi:10.1016/j.sbsr.2016.01.006.

[9] Yunusa, Z., Hamidon, M. N., Kaiser, A., & Awang, Z. (2014). Gas Sensors: A Review. Sensors & Transducers, 168(4), 61-75.

[10] Santoso, B., Wijayanto, H., Notodiputro, K. A., & Sartono, B. (2018). K-Neighbor over-sampling with cleaning data: a new approach to improve classification performance in data sets with class imbalance. Applied Mathematical Sciences, 12(10), 449–460. doi:10.12988/ams.2018.8231.

[11] Walimbe, R. (2017). Handling imbalanced dataset in supervised learning using family of smote algorithm. Data Science Central. Available online: https://www.datasciencecentral.com/handling-imbalanced-data-sets-in-supervised-learning-using-family/ (accessed on August 2022).

[12] Berthold, M. R., Cebron, N., Dill, F., Gabriel, T. R., Kötter, T., Meinl, T., Ohl, P., Thiel, K., & Wiswedel, B. (2009). KNIME - the Konstanz information miner. ACM SIGKDD Explorations Newsletter, 11(1), 26–31. doi:10.1145/1656274.1656280.

[13] Yıldırım, S. (2020). Gradient Boosted Decision Trees-Explained. Towards Data Science. Available online: https://towardsdatascience.com/gradient-boosted-decision-trees-explained-9259bd8205af (accessed on August 2022).

[14] Fabien, M. (2022). Gradient Boosting regression: Supervised Learning Algorithms. Switzerland. Available online: https://maelfabien.github.io/machinelearning/GradientBoost/ (accessed on August 2022).